\documentclass[letterpaper, 10 pt, conference]{ieeeconf}  % Comment this line out if you need a4paper

\usepackage[T1]{fontenc}
\usepackage[utf8]{inputenc}
\usepackage{authblk}
\usepackage{graphicx}
\usepackage[spaces,hyphens]{url}
\usepackage{array}
\usepackage{mathtools}
\usepackage{subcaption}
\usepackage{amssymb}
\usepackage{amsmath}
\usepackage{diagbox}
\usepackage{fourier} 
\usepackage{array}
\usepackage{makecell}
\usepackage{hyperref}
\usepackage{stackengine}
\usepackage{multirow}
\usepackage{sidecap}
\usepackage[flushleft]{threeparttable} % http://ctan.org/pkg/threeparttable
\usepackage{booktabs,caption}

\IEEEoverridecommandlockouts                              % This command is only needed if 
\title{\LARGE \bf
ROBI: A Multi-View Dataset for Reflective Objects in Robotic Bin-Picking
}

\author{Jun Yang$^{1}$, Yizhou Gao$^{2}$, Dong Li$^{2}$, and Steven L. Waslander$^{1}$
\thanks{$^{1}$Jun Yang and Steven L. Waslander are with University of Toronto Institute for Aerospace Studies and Robotics Institute.
        {\tt\footnotesize junyang.yang@mail.utoronto.ca, stevenw@utias.utoronto.ca}}
\thanks{$^{2} $Yizhou Gao and Dong Li are with Epson Canada
        {\tt\footnotesize \{yizhou.gao, dong.li\}@ea.epson.com}}
}

\begin{document}

\maketitle
\thispagestyle{empty}
\pagestyle{empty}

%%%%%%%%%%%%%%%%%%%%%%%%%%%%%%%%%%%%%%%%%%%%%%%%%%%%%%%%%%%%%%%%%%%%%%%%%%%%%%%%
\begin{abstract}
In robotic bin-picking applications, the perception of texture-less, highly reflective parts is a valuable but challenging task. The high glossiness can introduce fake edges in RGB images and inaccurate depth measurements especially in heavily cluttered bin scenario. In this paper, we present the ROBI (Reflective Objects in BIns) dataset, a public dataset for 6D object pose estimation and multi-view depth fusion in robotic bin-picking scenarios. The ROBI dataset includes a total of 63 bin-picking scenes captured with two active stereo camera: a high-cost Ensenso sensor and a low-cost RealSense sensor. For each scene, the monochrome/RGB images and depth maps are captured from sampled view spheres around the scene, and are annotated with accurate 6D poses of visible objects and an associated visibility score. For evaluating the performance of depth fusion, we captured the ground truth depth maps by high-cost Ensenso camera with objects coated in anti-reflective scanning spray. To show the utility of the dataset, we evaluated the representative algorithms of 6D object pose estimation and multi-view depth fusion on the full dataset. Evaluation results demonstrate the difficulty of highly reflective objects, especially in difficult cases due to the degradation of depth data quality, severe occlusions and cluttered scene. The ROBI dataset is available online at \url{https://www.trailab.utias.utoronto.ca/robi}.
\end{abstract}

%%%%%%%%%%%%%%%%%%%%%%%%%%%%%%%%%%%%%%%%%%%%%%%%%%%%%%%%%%%%%%%%%%%%%%%%%%%%%%%%
\section{INTRODUCTION}
Highly reflective objects are common in robotic bin-picking applications, and the goal is to have a vision-guided robot to pick up such objects with random poses from a bin. Towards this goal, highly accurate 6D object poses are often required prior to robot grasp execution. These reflective objects are mostly texture-less and cannot be reliably recognized with classic techniques that rely on local descriptors \cite{collet2011moped}. Instead, recent approaches that can deal with texture-less object have focused on depth or RGB-D based features \cite{hinterstoisser2012model,drost2010model,xiang2017posecnn,he2020pvn3d,rad2017bb8, doumanoglou2016recovering, sundermeyer2018implicit}. With the increasing availability of depth cameras, it brings the hope that object detection and 6D pose estimation can be eventually solved for such objects. As a result, RGBD-based object detection and pose estimation is an active research area for robotic bin-picking, and the RGBD datasets, consisting of RGB/monochrome image and depth maps, play an important role.

\begin{figure}[t]
\begin{subfigure}{.24\textwidth}
  \includegraphics[width=\linewidth]{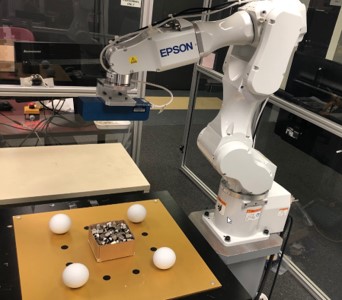}
  \vspace{-1.4\baselineskip}
  \caption{}
\label{fig1a}
\end{subfigure}
\begin{subfigure}{.24\textwidth}
  \includegraphics[width=\linewidth]{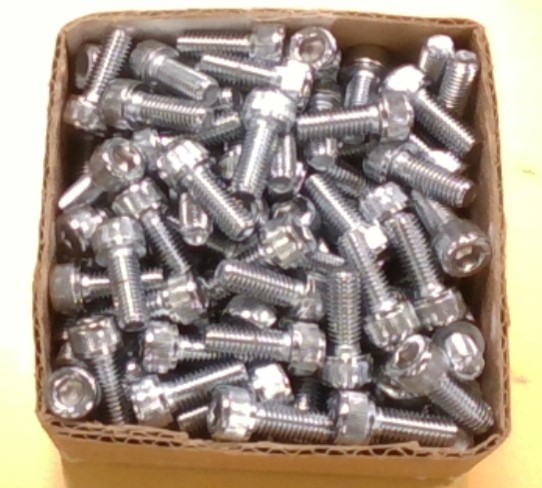}  
  \vspace{-1.4\baselineskip}
  \caption{}
\end{subfigure}
\begin{subfigure}{.24\textwidth}
  \includegraphics[width=\linewidth]{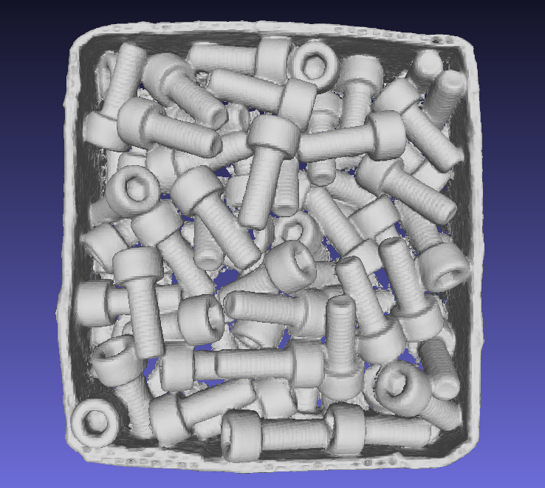}  
  \vspace{-1.4\baselineskip}
  \caption{}
\end{subfigure}
\begin{subfigure}{.24\textwidth}
  \includegraphics[width=\linewidth]{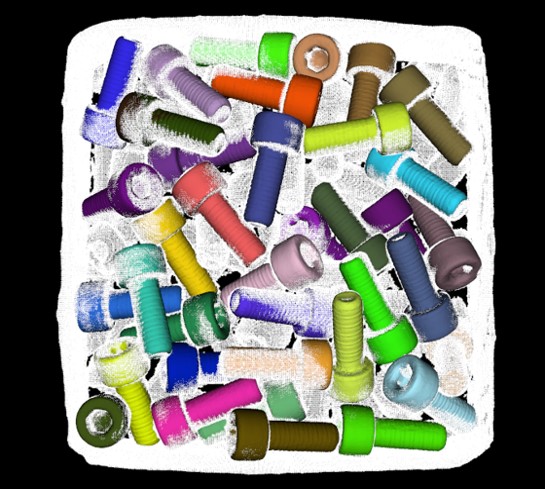}  
  \vspace{-1.4\baselineskip}
  \caption{}
\end{subfigure}
\caption{(a) The robot cell employed for data collection. (b) Bin-picking scene: objects are piled randomly in a bin. (c) Reconstructed 3D model of the scene. (d) The object models are rendered on top of the reconstructed scene with annotated ground truth poses.}
\label{fig1}
\end{figure}

\begin{figure*}[t]
\centering
  \includegraphics[width=\linewidth]{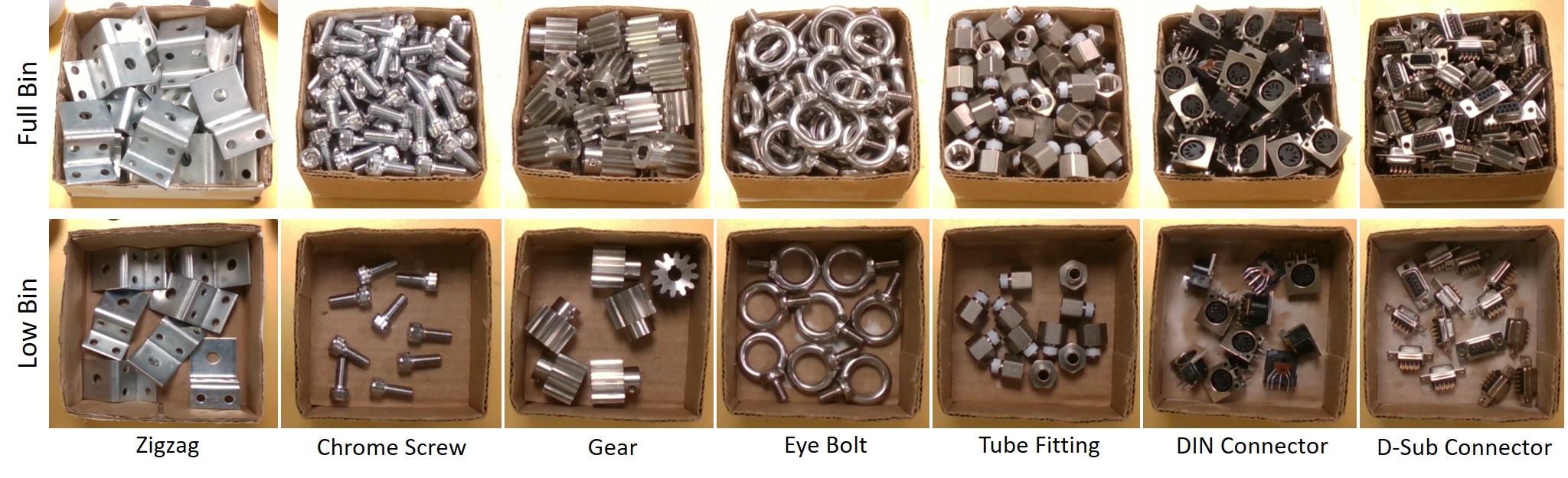}  
  \vspace{-1.4\baselineskip}
\caption{An overview of the dataset, including seven industrial objects, with two different bin scenarios.}
\label{fig2}
\end{figure*}

\begin{table*}[]
\centering
\resizebox{0.98\textwidth}{!}
{
\begin{threeparttable}
\begin{tabular}{|c||c|c|c|c|c|}
\hline
\textbf{Object} & \textbf{Diameter (mm)} & \textbf{Surface Material} & \textbf{Geometric Symmetry} & \textbf{Symmetry Order} & \textbf{Size Ratio}\tnote{$\dagger$}  \\ \hline
Zigzag          & 76.2                   & metallic                  & no                & -                       & 0.58            \\ \hline
Chrome Screw    & 29.1                   & metallic                  & yes               & infinite                & 0.46            \\ \hline
Gear            & 36.8                   & metallic                  & almost\tnote{*}             & 12                      & 0.89           \\ \hline
Eye Bolt        & 47.2                   & metallic                  & yes               & 2                       & 0.32            \\ \hline
Tube Fitting    & 24.5                   & metallic                  & yes               & 6                       & 0.81            \\ \hline
DIN Connector   & 32.5                   & metallic  and plastic     & no                & -                       & 0.62           \\ \hline
D-Sub Connector & 30.2                   & metallic  and plastic     & no                & -                       & 0.41          \\ \hline
\end{tabular}
\begin{tablenotes}
  \item[*] Almost symmetric: the objects breaks the symmetry with only a small detail.
  \item[$\dagger$] Size ratio: the ratio of the smallest to the largest sides of an axis-aligned object bounding box \cite{drost2017introducing}.
  \end{tablenotes}
  \end{threeparttable}
}
\caption{The object list in ROBI dataset with their properties.}
\label{tab1}
\end{table*}

However, industrial-grade depth cameras often fail to sense complete depths throughout the field of view when surfaces are too glossy, dark, close, or far from the sensor. Moreover, in the industrial environment, the severe clutter and occlusions of the scene can lead to significant degradation of depth quality. To this end, in our previous work \cite{fusion2021}, we constructed ROBI v1.0 dataset, a novel challenging dataset for evaluating different methods of multi-view depth fusion and 6D object pose estimation for highly reflective objects. It comprises 7 metallic industrial objects with different levels of reflectivity, and 5 individual bin instances with up to 88 viewpoints for each object. However, ROBI v1.0 only includes captures by a high-cost Ensenso active stereo camera, which is normally too expensive for the customer-level bin-picking system. To investigate lost-cost sensor performance on these reflective parts, in this paper, we expand the ROBI dataset to include more scenes from a low-cost Intel RealSense sensor. Our full dataset comprises 4 additional captures for each object, resulting in a total of 63 scenes and $\sim$8K images. The capture platform is equipped with two active stereo camera sensors, a high-cost Ensenso N35, and a low-cost RealSense D415. In addition, to study the impact of depth quality for different perception solutions, we provide ground truth depth maps captured by the high-cost Ensenso with objects coated in anti-reflective scanning spray.

We use a robot manipulator to move the sensors to different viewpoints, systematically sampling from a spherical dome. We separate our data capture into full-bin and low-bin scenarios to demonstrate two different bin-picking conditions, as shown in Fig. \ref{fig2}. For each scene and viewpoint, monochrome/RGB images, raw and ground truth depth maps, and annotations of 6D object poses with visibility scores are included.

The ROBI dataset is intended for evaluating different perception solutions for highly reflective objects, such as 6D object pose estimation \cite{hodan2018bop}, scene reconstruction \cite{fusion2021}, depth completion \cite{sajjan2020clear}, and active perception \cite{sock2019active}. We capture images with two different grades of sensors, associated with ground truth depth maps. This allows us to study the impact of different input modalities and depth data quality for a given problem. The difficulty of reflective objects for pose estimation and multi-view depth fusion is demonstrated by the relatively low performance using representative algorithms in Sec. \ref{sec4}.

In summary, we make the following contributions:
\begin{itemize}
	\item A real-world multi-view dataset for reflective objects in the bin-picking scenarios with: {1)} 7 metallic industrial parts with different levels of reflectivity, {2)} 63 bin instances for each object with up to 125 distinct viewpoints, {3)} images captured with a high-cost Ensenso camera and a low-cost RealSense camera.
	\item A novel method to acquire the ground truth depth maps for highly reflective objects, which are missing in most of datasets.
	\item The evaluation of the representative approaches for object pose estimation and multi-view depth fusion on the full ROBI dataset.
\end{itemize}

\section{RELATED WORK}
\label{sec2}
The progress of research in computer vision and robotics has been strongly influenced by datasets, which enable us to evaluate methods and understand their limitations. In this section, we review related datasets for robotic bin-picking.

\subsection{Datasets for Object Detection and 6D Pose Estimation.}
Object detection and 6D pose estimation play an important role in many technological areas. With the increasing availability of RGB-D cameras, numerous datasets has appeared \cite{hinterstoisser2012model,hodan2017t,tejani2014latent, xiang2017posecnn, kaskman2019homebreweddb,kleeberger2019large,drost2017introducing}. A summary of all these datasets is presented in \cite{kleeberger2019large}, and in the benchmark for 6D object pose estimation (BOP) \cite{hodan2018bop}, the authors have performed a comprehensive evaluation of 15 diverse approaches on eight recent datasets.

The LINEMOD Dataset \cite{hinterstoisser2012model} is a widely used benchmark for texture-less objects. It contains $\sim$18000 RGB-D images of 15 objects, and has become a standard benchmark in most of recent works \cite{hinterstoisser2012model,drost2010model,xiang2017posecnn,he2020pvn3d}. The test images feature severe clutter but only mild occlusion. This work was augmented by \cite{brachmann2014learning} to consider a high degree of occlusion for evaluation. Datasets presented in \cite{tejani2014latent, xiang2017posecnn, kaskman2019homebreweddb} have similar properties. The objects have discriminative color, large size, and limited pose variability, making recognition relatively easy.

For industrial object detection, Drost et al. \cite{drost2017introducing} introduced the ITODD dataset, which contains 28 industrial parts with minor reflective surfaces. The T-LESS Dataset \cite{hodan2017t} features 30 texture-less objects with no discriminative color and shape. The test images were captured from 20 scenes with various complexity: from several isolated objects on a clean background to multiple objects stacked with severe occlusions and clutter. Despite numerous advantages, these two datasets have the following limitations: \textbf{1)}: they lack pose and scene variations, hence cannot represent industrial warehouse scenarios; \textbf{2)} all objects have low reflectivity and do not encounter adverse effects of noisy and missing depth measurements, which is common to industrial parts. 

To better serve the bin-picking scenario, Doumanoglou et al. \cite{doumanoglou2016recovering} provided the IC-BIN dataset with multiple objects stacked in a bin. It compromises three scenes of two objects from IC-MI dataset \cite{tejani2014latent}. The same scene was recorded from different viewpoints for evaluating object pose estimation and active vision techniques. Recently, the Fraunhofer IPA dataset \cite{kleeberger2019large} has been introduced for object pose estimation and instance segmentation for robotic bin-picking. However, it contains only 520 real-world depth images of two industrial objects. In comparison, we captured a total of $\sim$8K images for 7 objects in the real-world.

\subsection{Datasets for Multi-View Depth Fusion.}
Historically, 6D object pose estimation has been addressed from a static viewpoint. However, depth cameras often produce inaccurate depth measurements or fail to sense depths entirely from a single viewpoint due to reflective object materials, limited sensor resolution, and scene occlusions. To overcome these limitations, when setup permits, multi-view fusion approaches are able to provide higher levels of scene completion than single view acquisition.

There are several datasets of multi-view RGB-D scans, including ScanNet \cite{dai2017scannet}, Matterport3D \cite{chang2017matterport3d}, and others \cite{xiao2013sun3d, zhou2013dense, choi2016large}. However, there is no ground-truth available for qualitative evaluation of depth fusion or scene reconstruction. In \cite{weder2020routedfusion, riegler2017octnetfusion}, authors obtained the ground truth by fusing all frames of each scene using standard TSDF fusion \cite{curless1996volumetric} and manually removed outliers. The evaluation was then performed on only a subset of frames. Compared to these methods, we provide a novel acquisition method to acquire high-quality ground truth depth images for multi-view depth fusion and scene reconstruction.

\section{ROBI DATASET}
\label{sec3}
Compared to the existing datasets, the ROBI dataset has three unique characteristics. First, it contains 7 highly reflective objects, with different geometric properties that are common in industrial bin picking, but are not represented existing datasets due to the difficulty of working with specular reflection. Second, we captured different bin scenarios using two different grades of cameras with multi-view data acquisition. This allows us to test the impact of different data modality, viewpoint variation and clutter of the scene. Finally, we provided ground truth depth maps captured by the high-cost depth camera with parts coated in anti-reflective scanning spray. This allows us to quantitatively study the impact of input depth quality for different robotic perception tasks. To the best of our knowledge, this is the first dataset for highly reflective objects with ground truth depth maps. In this section, we describe the details of dataset creation, including sensor setup, data capture pipeline, object model generation, and ground truth annotation.

\begin{figure}[t]
\centering
\begin{subfigure}{0.25\textwidth}
  \includegraphics[width=\linewidth]{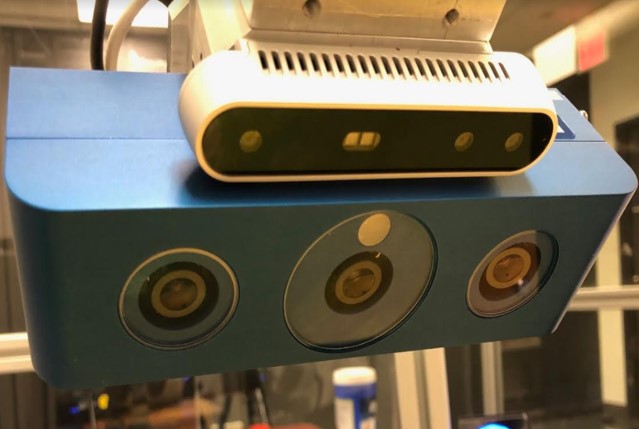}
  \vspace{-1.4\baselineskip}
  \caption{}
  \label{fig10a}
\end{subfigure}
\begin{subfigure}{0.20\textwidth}
    \includegraphics[width=\linewidth]{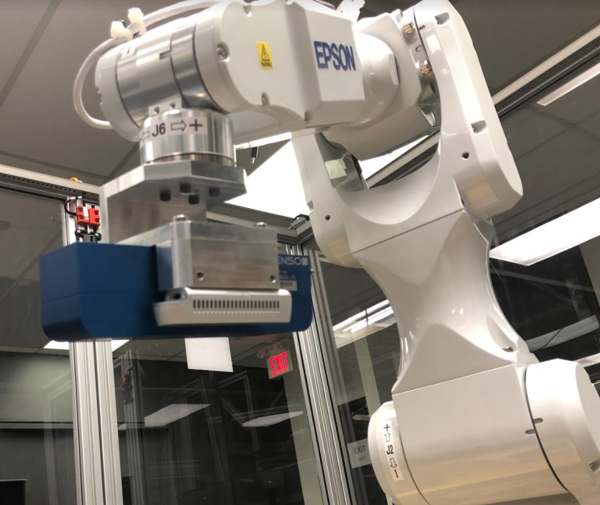}
    \vspace{-1.4\baselineskip}
    \caption{}
    \label{fig10b}
\end{subfigure}
\caption{(a) The two active stereo cameras used for data collection. (b) we mount the Ensenso and RealSense on an EPSON C4L robot arm.}
\label{fig10}
\end{figure}

\subsection{Sensor Setup}
Due to the requirements of high accuracy and short cycle time in robotic bin-picking, we used active stereo cameras for our data capture. Specifically, as shown in Fig. \ref{fig10a}, a high-cost Ensenso N35 depth sensor and a low-cost Intel RealSense D415 RGB-D camera were used for our real-world data capture. We calibrated intrinsic parameters for both cameras with a precisely manufactured calibration board (shown in Fig. \ref{fig1a}) and calibration toolboxes provided by camera vendor. The root mean square re-projection errors calculated on circles of the calibration board are $0.22\;px$ for Ensenso, and $0.12\;px$ and $0.3\;px$ for RealSense's RGB and depth frames, respectively.

\begin{figure}[t]
\centering
\begin{subfigure}{0.48\textwidth}
  \includegraphics[width=\linewidth]{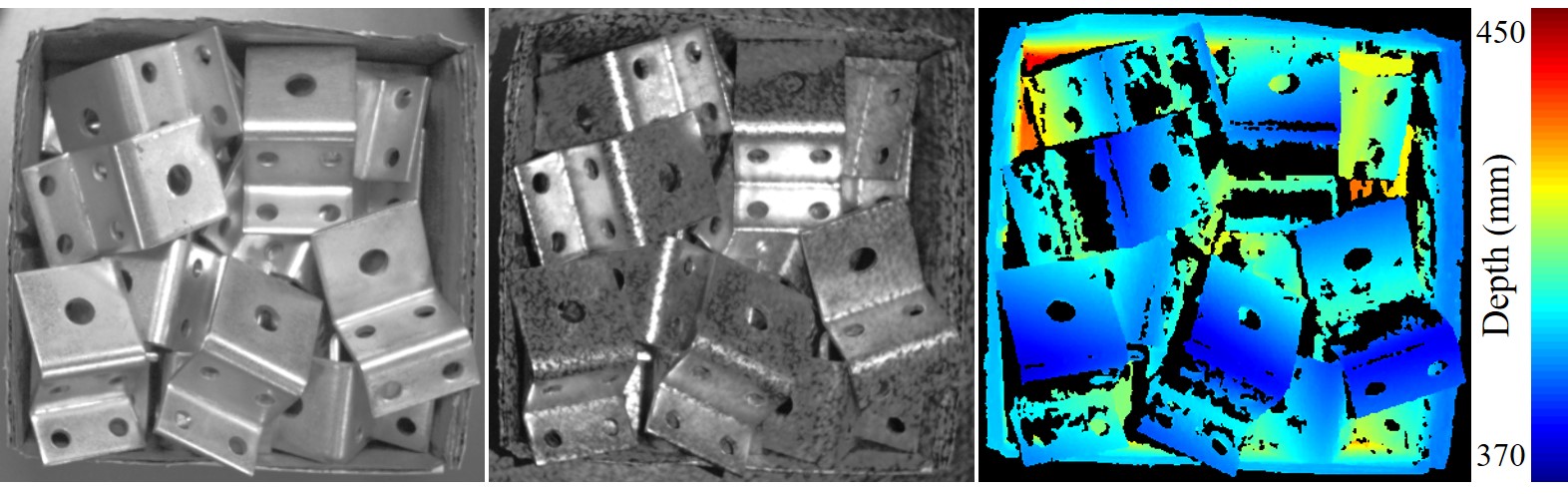}
  \vspace{-1.4\baselineskip}
  \caption{Ensenso N35 camera data}
  \label{fig3a}
\end{subfigure}
\begin{subfigure}{0.48\textwidth}
    \includegraphics[width=\linewidth]{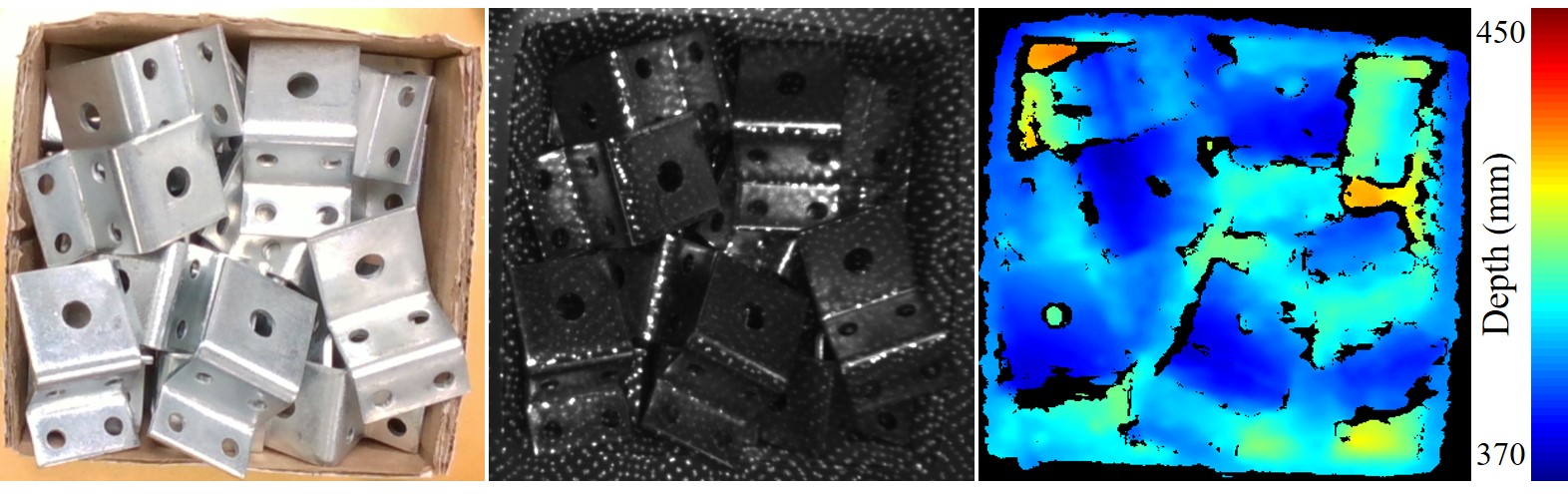}
    \vspace{-1.4\baselineskip}
    \caption{RealSense D415 camera data}
    \label{fig3b}
\end{subfigure}
\caption{Camera data for Ensenso N35 and RealSense D415. From left to right: RGB/monochrome image, patterned projected image, and depth map.}
\label{fig3}
\end{figure}

The Ensenso N35 camera comes with an optical lens for short working distance. It has a minimum working distance of $240\;mm$, a maximum working distance of $520\;mm$, and an optimal working distance of $330\;mm$. The sensor is comprised of two visible-light cameras and a light projector, and produces depth maps with a resolution of $1280\times1024$. Since RGB images are not available, we captured Ensenso data with the following two configurations:
\begin{itemize}
\item \textbf{Depth configuration.} We turn the camera projector on and use a low exposure time to eliminate the impact of ambient light. The raw disparity maps and pattern projected stereo pairs are collected.
\item \textbf{Monochrome configuration.} The camera projector is turned off for this configuration to capture monochrome images. We use a higher exposure time to obtain optimal contrast for objects. Only stereo pair data is saved with this configuration.
\end{itemize}
Fig. \ref{fig3a} shows the sample camera data captured with Ensenso N35 for the "ZIGZAG" object . The depth map demonstrates that the Ensenso is able to capture fine geometric details, but there is a significant amount of missing data due to surface reflection.

The Intel RealSense D415 is a compact, low cost sensor. It comes with an RGB camera and a depth camera system, comprising two IR cameras and an IR projector. We used the RealSense camera to capture the synchronized RGB images, depth maps, and pattern projected stereo pairs at $1280\times720$ resolution. The working distance is tuned to the range of $[243\,mm, 638\,mm]$. As shown in Fig. \ref{fig3b}, compared to Ensenso, the depth map captured with RealSense has a higher level of completeness, but with significant degradation of depth accuracy (e.g., loss of geometric details).

\subsection{Data Capture Pipeline}
\begin{figure}[t]
\begin{subfigure}{.255\textwidth}
  \includegraphics[width=\linewidth]{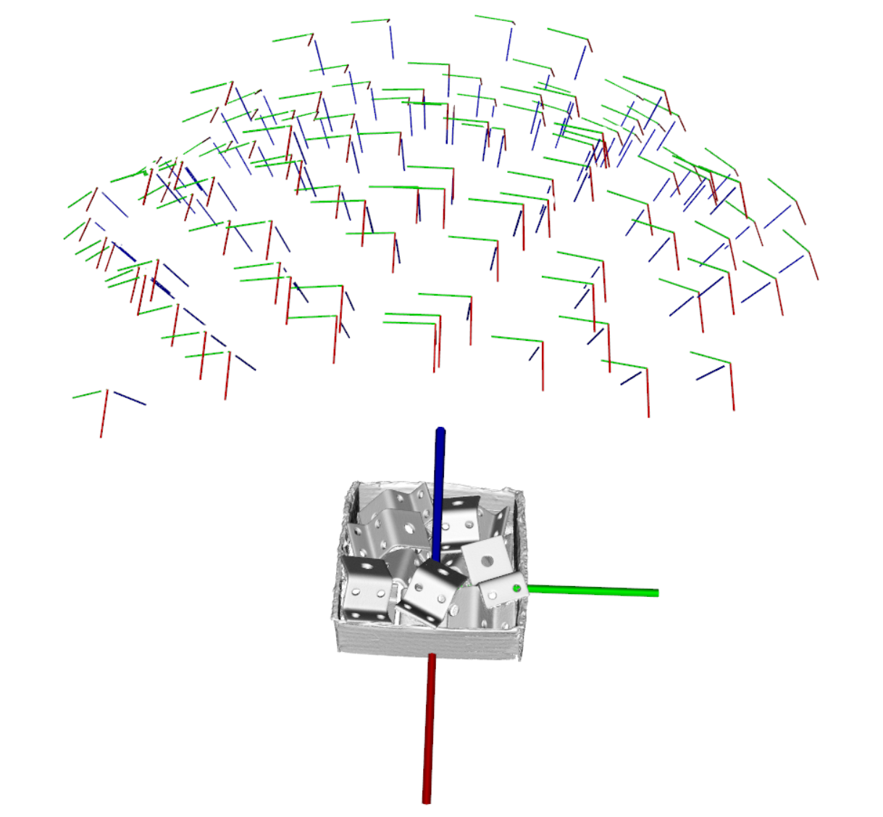}
  \vspace{-1.4\baselineskip}
  \caption{Full bin}
\end{subfigure}
\begin{subfigure}{.225\textwidth}
  \includegraphics[width=\linewidth]{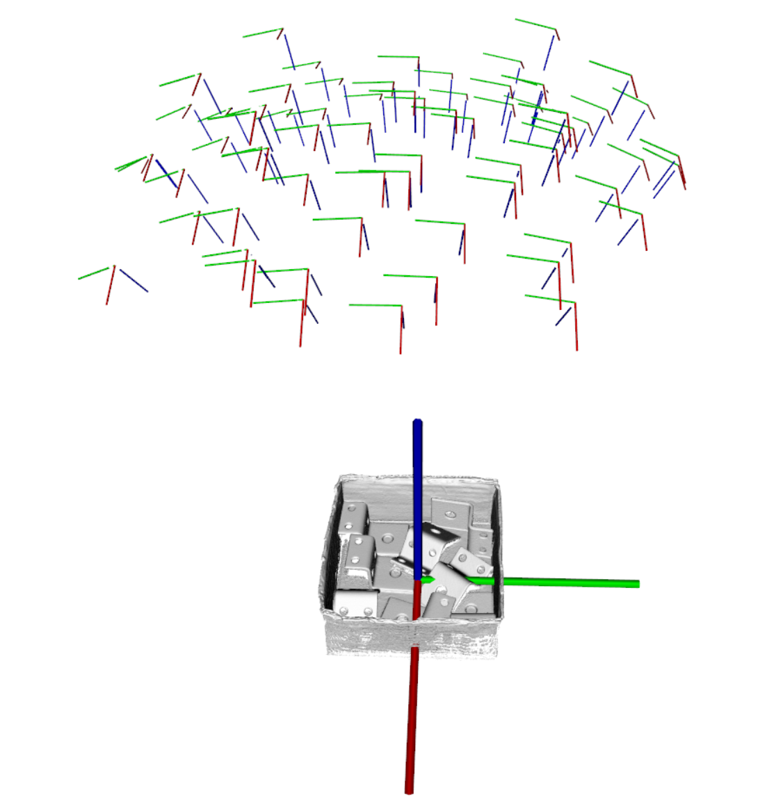}  
  \vspace{-1.4\baselineskip}
  \caption{Low bin}
\end{subfigure}
\caption{Sampled viewpoints for multi-view data capture using Ensenso camera.}
\label{fig4}
\end{figure}

For each of the seven objects depicted in Fig \ref{fig2}, we captured four bin scenarios (2 full-bin and 2 low-bin). The two active stereo cameras are mounted to an EPSON 6-Axis C4L robot arm, illustrated in Fig. \ref{fig10b}. We program the robot arm to move on a view sphere around the scene, from approximately $45^{\circ}$ to $90^{\circ}$ of elevation with different camera distances. The robot end-effector stays pointing towards the center of the workstation. For full-bin scenarios, we sampled a total of 106 views for Ensenso and 125 views for RealSense. Due to the occlusion of bin wall, in low-bin data capture, the sphere elevation is limited to the range of $[65^{\circ}, 90^{\circ}]$, and there are 68 and 94 views for Ensenso and RealSense, respectively. Fig. \ref{fig4} demonstrates the sampled viewpoints for these two different scenes, captured with Ensenso camera.

To determine 6D camera poses at different viewpoints, we use a two-step calibration procedure: {1)} initial camera pose estimation using robot forward kinematics and the calibrated poses between end-effector and cameras; {2)} pose refinement by performing iterative closest point (ICP) on calibration spheres. To acquire the transformation between robot end-effector and two cameras, we leveraged the industrial-grade accuracy of our C4L robot arm, and move the cameras to look at the same calibration pattern at different viewpoints. The 6D poses for Ensenso and RealSense camera are then estimated by solving the classic "hand-eye" calibration problem \cite{daniilidis1999hand}. For better refinement accuracy using ICP algorithm, we placed calibration spheres around the bin (illustrated in Fig. \ref{fig1a}). The spheres are manufactured with a matte surface, so that the cameras can achieve their optimal accuracy. The average closest-point residual error was successfully reduced from 0.33 mm to 0.26mm for Ensenso data, and 1.52 mm to 0.8 mm for RealSense data.

\begin{figure}[t]
\begin{subfigure}{.48\textwidth}
  \includegraphics[width=\linewidth]{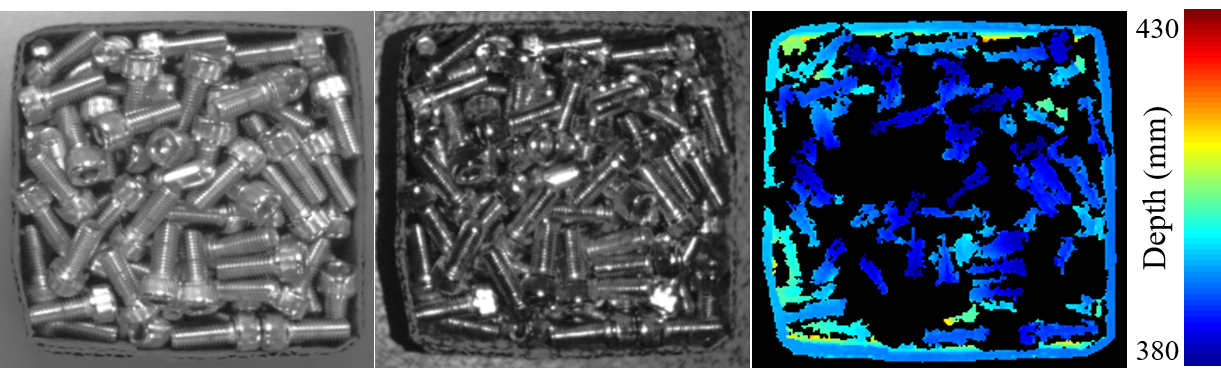}
  \vspace{-1.4\baselineskip}
  \caption{Ensenso data without scanning spray.}
  \label{fig5a}
\end{subfigure}
\begin{subfigure}{.48\textwidth}
  \includegraphics[width=\linewidth]{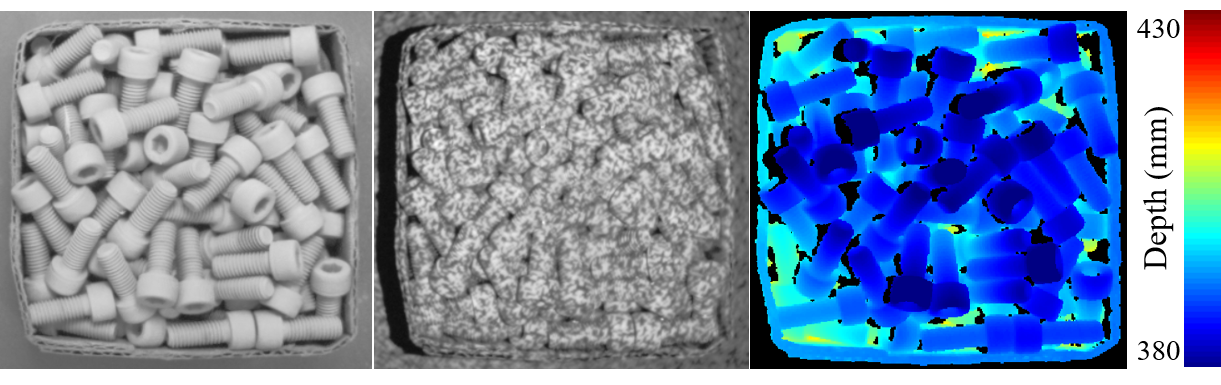}  
  \vspace{-1.4\baselineskip}
  \caption{Ensenso data with scanning spray.}
  \label{fig5b}
\end{subfigure}
\begin{subfigure}{.48\textwidth}
  \includegraphics[width=\linewidth]{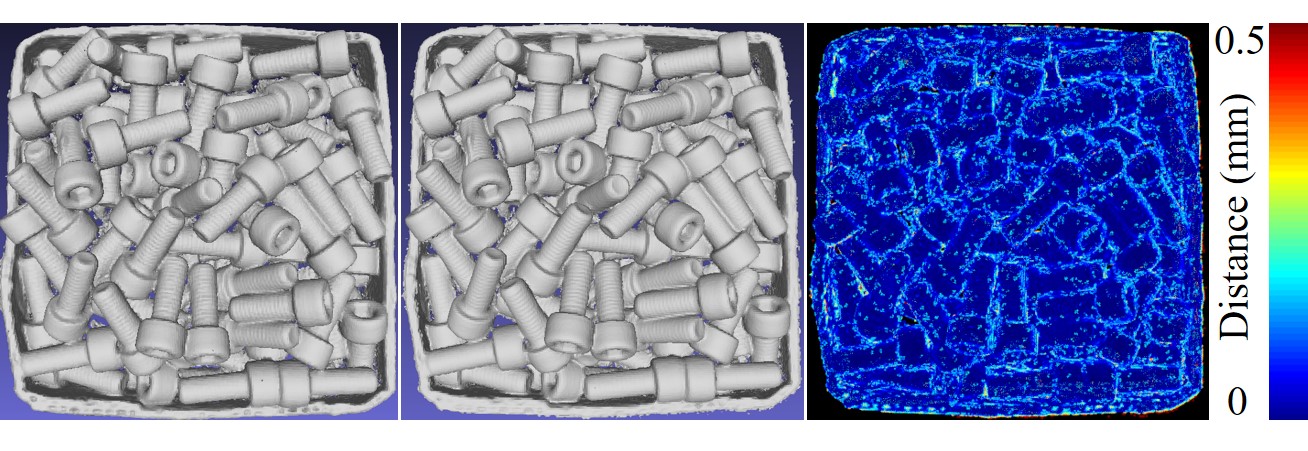}  
  \vspace{-1.4\baselineskip}
  \caption{Reconstructed scene model.}
  \label{fig5c}
\end{subfigure}
\caption{(a)$\&$(b) \textbf{Illustration of using scanning spray.} From left to right: monochrome images, pattern projected images, and depth maps. The ground truth mesh using TSDF fusion. (c) \textbf{Reconstructed scene model.} From left to right: Reconstructed scene using TSDF fusion, probabilistic fusion \cite{fusion2021}, and heatmap of the point-to-point distance between these two methods.}
\label{fig5}
\end{figure}

\begin{figure}[t]
\centering
  \includegraphics[width=\linewidth]{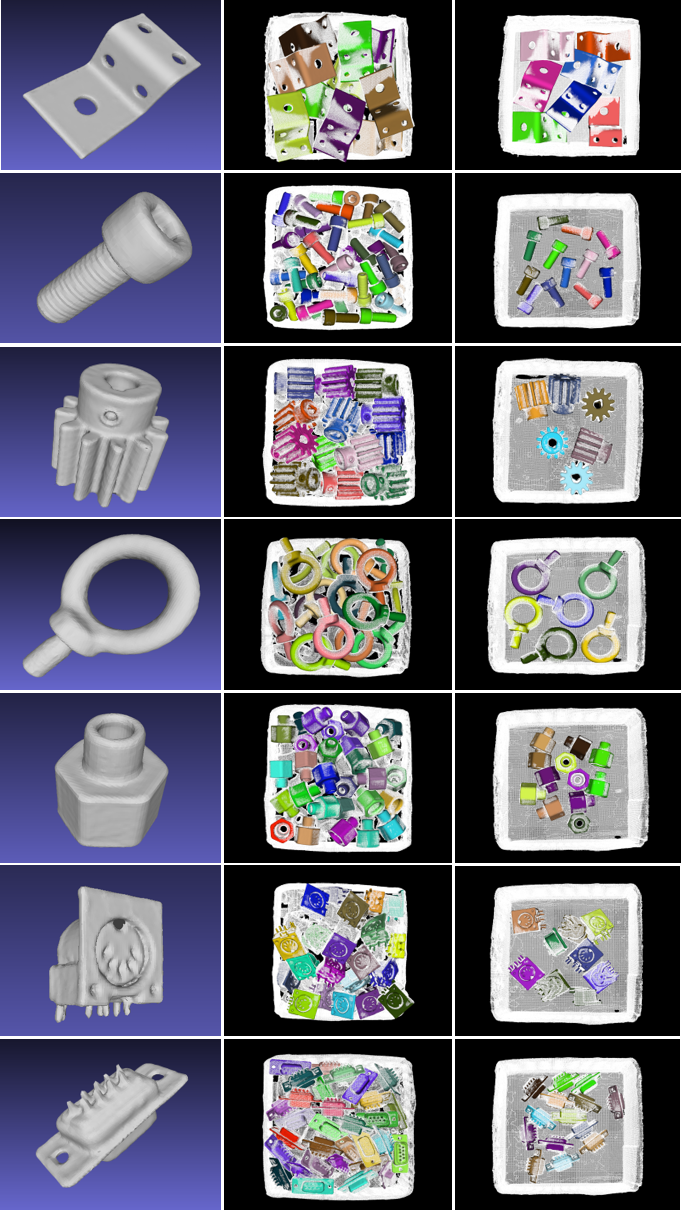}  
  \vspace{-1.4\baselineskip}
\caption{Object models and visualization of ground truth poses in ROBI dataset. From left to right in each row: reconstructed object model, examples of full-bin and low-bin scene, overlaid with the ground truth poses.}
\label{fig6}
\end{figure}

\subsection{Ground Truth Depth Maps and Scene Model}
\label{sec3c}
Active stereo cameras can provide reliable depth measurements when object surfaces have the ideal diffuse reflection, also known as Lambertian reflectance. However, most industrial objects do not have this property, resulting in inaccurate or missing depth measurements. This phenomenon is more obvious for objects with glossy surfaces. As demonstrated in Fig. \ref{fig5a}, due to severe specular and inter-reflections in the bin, the high-cost Ensenso camera fails to sense a large amount of depth data. To this end, some techniques have been studied to recover missing depth, such as depth completion \cite{sajjan2020clear} and multi-view depth fusion \cite{fusion2021} for the bin-picking problem. In this section, we provide a method to capture the ground truth depth maps in the real world.

We captured the ground truth depth maps with the Ensenso camera, and applied a scanning spray \cite{AESUB} on objects to better approach ideal Lambertian surfaces, so that the Ensenso camera can achieve its optimal accuracy ($\sim 0.2 mm$). The scanning spray generates a homogeneous layer with only $8-15\,\mu m$ thickness, which is at least one order of magnitude less than expected depth accuracy. The spray is able to self-evaporate within a few hours. Therefore, we leveraged the high repeatability of our robot arm and captured the test and ground truth images with two separate scans. In the first scan, we applied the scanning spray and capture ground truth depth maps. After full evaporation, the test images were captured during the second scan. Fig. \ref{fig5b} shows the captured images after applying the scanning spray. Compared with images without the spray (Fig. \ref{fig5a}), the Ensenso sensor is able to capture more complete and less noisy depth maps on the same scene. 

\begin{figure*}[t]
\centering
\begin{subfigure}{0.98\textwidth}
  \includegraphics[width=\linewidth]{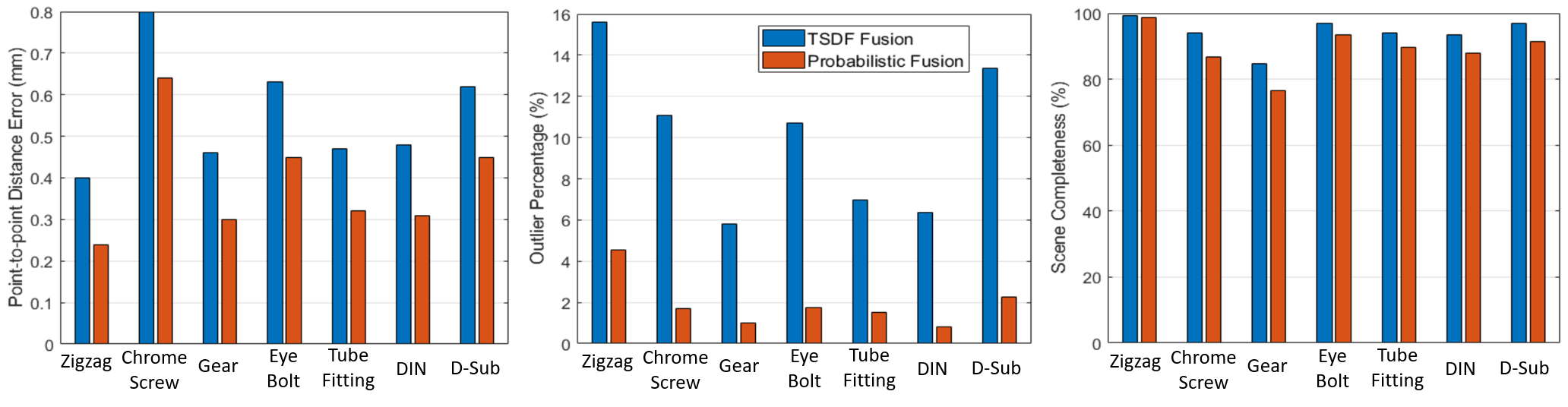}
  \caption{Quantitative reconstruction results on Ensenso camera data}
\end{subfigure}
\begin{subfigure}{0.98\textwidth}
    \includegraphics[width=\linewidth]{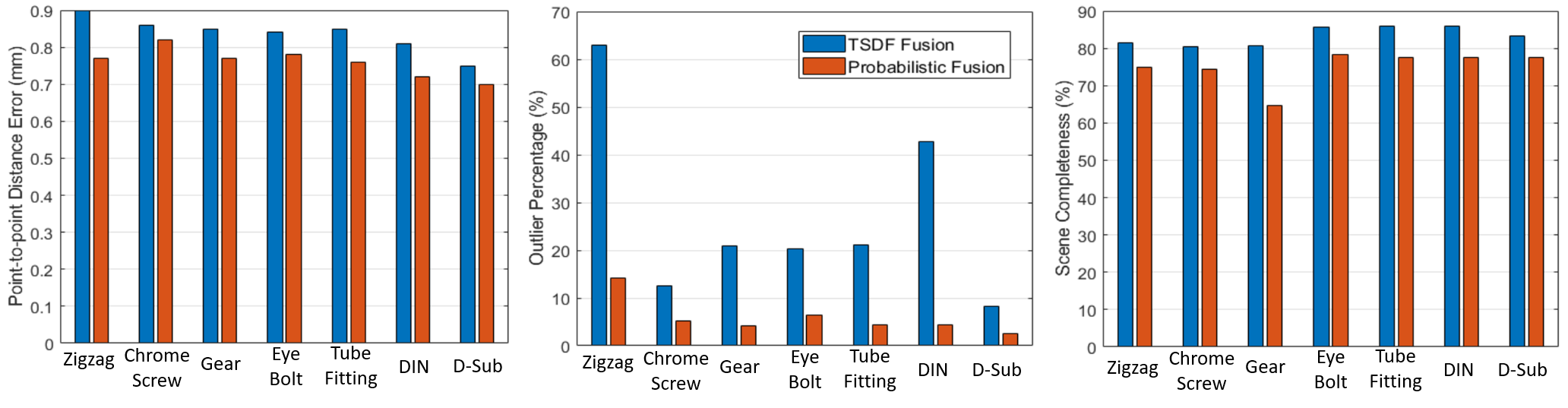}
    \caption{Quantitative reconstruction results on RealSense camera data}
\end{subfigure}
\caption{(a) Quantitative evaluation results for multi-view volumetric depth fusion, regarding mean point-to-point distance, outlier percentage, and scene completeness for all the objects in the full ROBI dataset.}
\label{fig7}
\end{figure*}

To evaluate depth fusion methods, for each scene, we constructed a ground truth mesh by applying TSDF fusion {\cite{newcombe2011kinectfusion}} on ground truth depth maps. To demonstrate that the reconstruction of the ground truth scene is not biased to any fusion method, we apply two methods, TSDF fusion and probabilistic fusion \cite{fusion2021} to produce two sets of ground truth meshes, as shown in Fig. \ref{fig5a} and \ref{fig5b}. We compute the mean point-to-point distance to quantitatively measure the difference between these two sets. The distance is less than 0.03 mm, indicating that our ground truth accuracy is not sensitive to different fusion methods.

\subsection{3D Object Model}
For each object, we provide both a manually created CAD model and a semi-automatically reconstructed model. Both models are provided with the dataset in the form of 3D meshes.

The reconstructed model is created in a manner similar to the creation of the scene model. We capture the multi-view high-quality depth maps after applying scanning spray, and use TSDF fusion for model reconstruction. For each object, we first reconstruct 2-4 partial models (depending on its geometric properties), and manually remove unnecessary parts and minor artifacts for each partial model. The partial models were then registered into a global reference frame by manual alignment and ICP refinement. Finally, we used MeshLab \cite{scarano2008meshlab} to inpaint minor holes in the model and smooth the mesh surface. The reconstructed models are shown in Fig. \ref{fig6} (first column).

\subsection{Ground Truth Poses}
We provide ground truth object poses using the CAD object model and the high-quality scene model, described in Sec. \label{sec3c}. We perform the alignment process using MeshLab \cite{scarano2008meshlab}. We first manually aligned object models to the scene model as an initial 6D object pose estimate. We then upsample the CAD model to a high resolution, and apply ICP to refine the object pose. Lastly, we adjust the pose manually whenever necessary to correct any in-plane translation and rotation. We repeat this process several times until a satisfactory alignment is achieved.

Lastly, for each ground truth pose, we generate a visibility score in the scene. This is done by rendering the segmentation mask of object pose via perspective projection. The visibility score, $\mathbf{v} \in [0, 1]$, is the fraction of the number of visible pixels over the number of pixels without any occlusion.

\begin{figure*}[t]
\begin{subfigure}{.49\textwidth}
  \includegraphics[width=\linewidth]{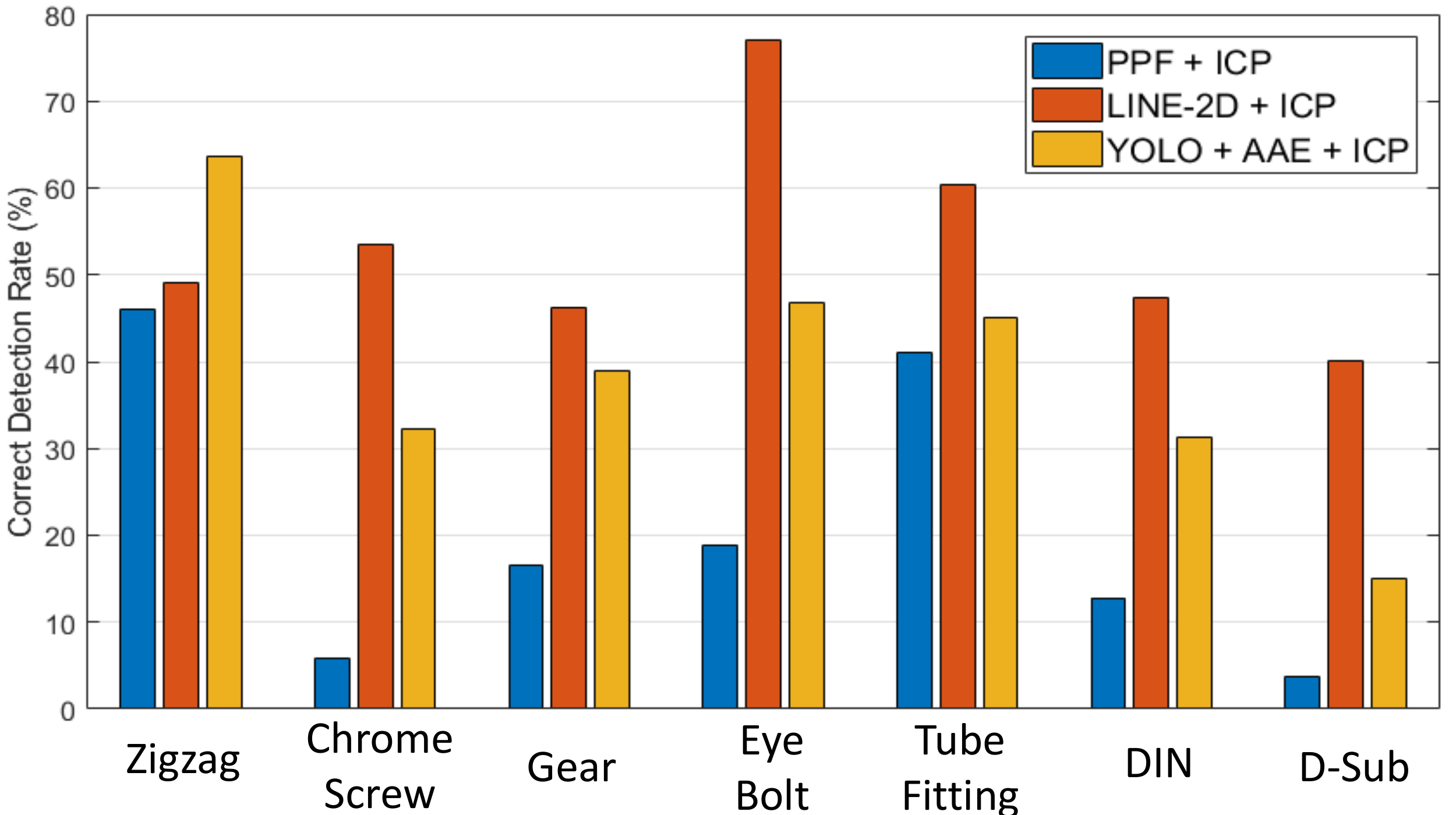}
  \vspace{-1.4\baselineskip}
  \caption{Ensenso camera data}
\end{subfigure}
\begin{subfigure}{.49\textwidth}
  \includegraphics[width=\linewidth]{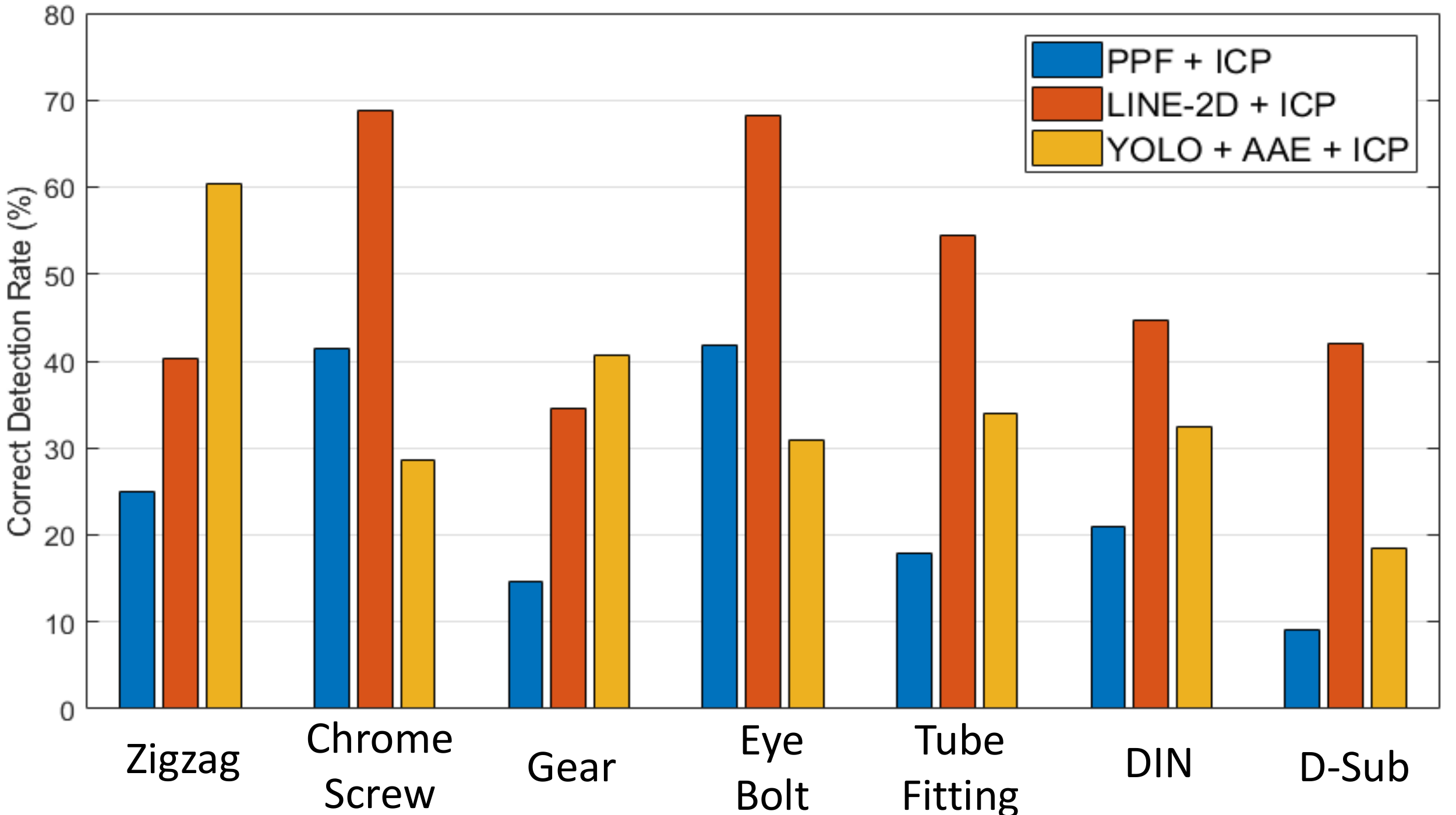}  
  \vspace{-1.4\baselineskip}
  \caption{RealSense camera data}
\end{subfigure}
\caption{(a) Detection rate results for different objects in the full ROBI dataset, including both Ensenso and Realsense camera data.}
\label{fig8}
\end{figure*}

\begin{figure}[t]
  \includegraphics[width=\linewidth]{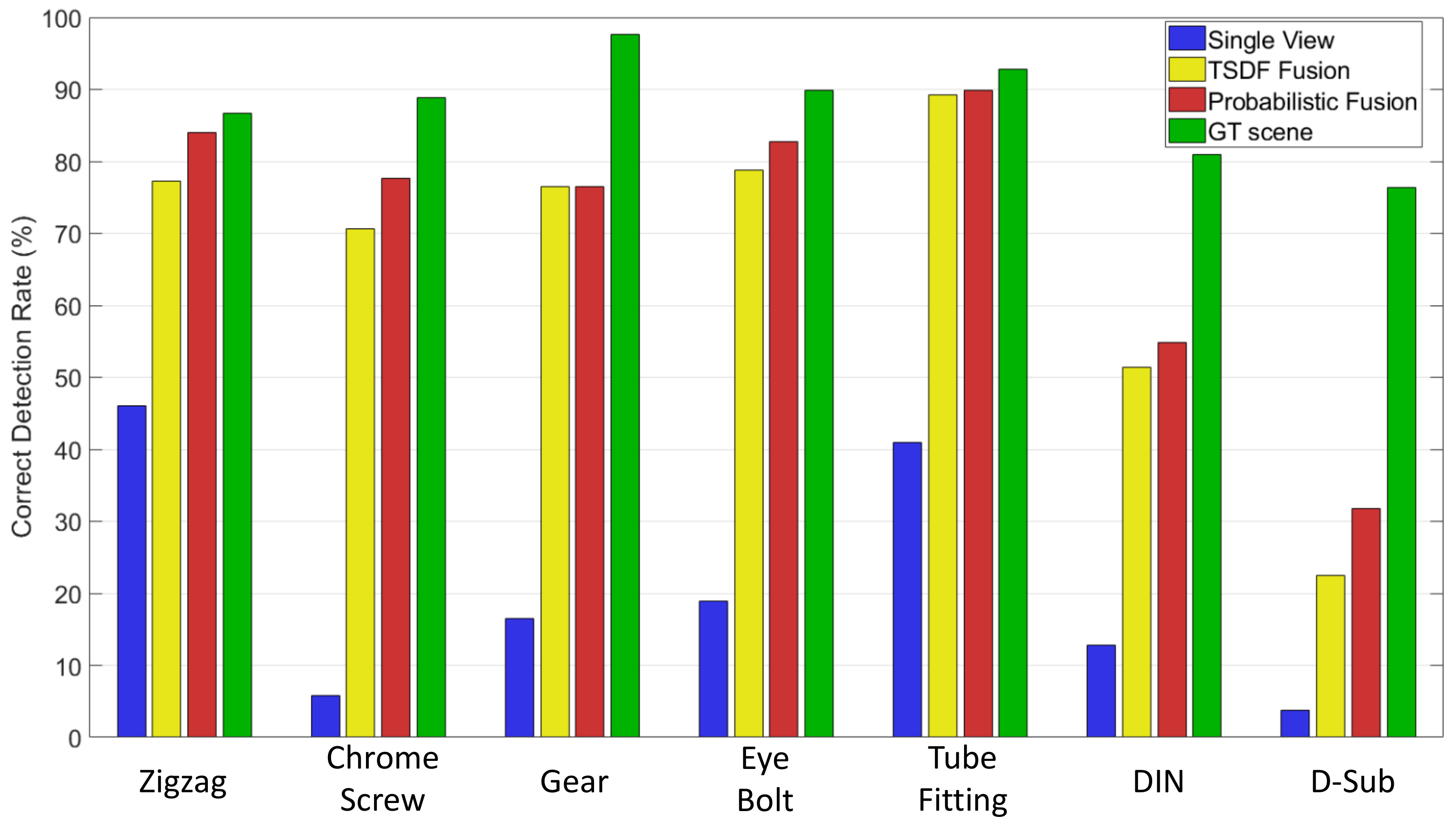}  
  \vspace{-1.4\baselineskip}
\caption{(a) Detection rate results of PPF pose estimator on Ensenso camera data, with different level of input depth data quality.}
\label{fig9}
\end{figure}

\section{EXPERIMENTS AND BENCHMARKS}
\label{sec4}
In this section, we present different benchmarks evaluating the performance of multi-view depth fusion and 6D object pose estimation on reflective objects. All experiments were performed on the full ROBI dataset.

\subsection{Multi-View Depth fusion}
\label{sec4a}
For the performance of depth fusion, we evaluated two volumetric fusion approaches, the traditional TSDF fusion \cite{curless1996volumetric} and a probabilistic fusion approach from \cite{fusion2021}. We use three metrics from \cite{fusion2021} for the evaluation: mean point-to-point distance, outlier percentage, and scene completeness. These metrics are performed only on the surface of objects, which has a major impact on object pose estimation.

Fig. \ref{fig7} shows the quantitative reconstruction results for each object with all viewpoints from different cameras. It can be seen that, for both Ensenso and RealSense data, the probabilistic fusion \cite{fusion2021} has a much lower reconstruction error than TSDF method on all objects. Moreover, it provides a better trade-off between scene completeness and outlier percentage due to its explicit uncertainty handling. However, we can observe that the probabilistic fusion sacrifices more completeness of scene on the "Gear" object, which has both a high-gloss surface and a complex geometric shape. Fig. \ref{fig7} also reveals that, when compared with Ensenso results, the reconstruction performance drops significantly on RealSense data, indicating that both methods are upper bounded by the quality of depth measurements.

\subsection{6D Object Pose Estimation}
To measure the 6D pose error of the object, we used the standard Average Distance of Model Points (ADD) for non-symmetric objects and ADD-S for symmetric objects~\cite{hinterstoisser2012model}. A pose hypothesis is accepted as the correct detection if its ADD or ADD-S score is less than 10\% of the object diameter.

For the performance evaluation, we chose three representative pose estimators: LINE-2D \cite{hinterstoisser2012model}, PPF (point pair feature) \cite{drost2010model} and AAE (augmented autoencoder) \cite{sundermeyer2018implicit}. These methods only require the object model for training. The LINE-2D and AAE are RGB-based pipelines. LINE-2D is a template-based method that builds the multi-view templates from the 3D models, and exploits the gradient response on monochrome/RGB images for detection in run-time \cite{hinterstoisser2011gradient}. AAE is a learning-based pose estimator. The authors used the self-supervised Augmented Autoencoders for estimating 3D rotation, and object detector (e.g., YOLO \cite{redmon2016you}) for translation estimation. Compared with these two approaches, the method proposed by Drost et al. \cite{drost2010model} relies on 3D point cloud data as input, and solves object poses by coupling the idea of point pair features and a dense voting scheme. This method is still one of the best performings on the 6D object pose estimation leader board (BOP) \cite{hodan2018bop}. We implemented these methods based on OpenCV \cite{bradski2000opencv}, Point Cloud Library (PCL) \cite{rusu20113d} and open-source codes provided by authors \cite{sundermeyer2018implicit}. The rendering of training images was based on a toolkit from BOP \cite{hodan2018bop}. Lastly, we apply the iterative closest point (ICP) algorithm for the final refinement of the object pose estimates.

As in \cite{hodan2017t, drost2010model, rad2017bb8}, the performance is measured by correct detection rate (CDR). A ground truth pose will be taken into consideration only if its visibility score is larger than 0.6. Fig. \ref{fig8} presents the CDR for each object using single view data. It can be seen that, due to a large number of missing depth measurements, the PPF method has an overall poor performance on both Ensenso and RealSense data. In comparison, the LINE-2D and AAE methods have significantly higher detection rates since they are less dependent on depth data (used for ICP only).

To improve the performance of pose estimation, as illustrated in \cite{fusion2021}, we can apply the depth fusion and provide high-quality depth input data for the pose estimator. We demonstrate this strategy on the Ensenso camera data with PPF pose estimator, whose performance is highly correlated to depth data. Fig. \ref{fig9} shows the results on two volumetric fusion approaches, evaluated in Sec. \ref{sec4a}, as well as the ground truth scene model. It can be seen that, compared with the single view result, the performance of PPF pose estimator can be greatly improved with multi-view depth fusion. And as a result of the higher reconstruction accuracy and few outliers, the probabilistic approach outperforms TSDF fusion on almost all objects. When given more accurate 3D data in the form of the ground truth meshes, the PPF pose estimator can provide close to perfect detection rates.

\section{CONCLUSION}
\label{sec5}
This paper has presented ROBI, a new dataset for evaluating various robotic perception tasks on reflective objects. The dataset features challenging, highly reflective objects and different real-world bin scenarios which were captured using different grades of cameras with multiple viewpoints. Further, to investigate the impact of depth data quality for a given problem, we provide the ground truth depth maps captured by a high-cost depth sensor with objects coated in anti-reflective scanning spray. Initial evaluation results using the ROBI dataset indicate that the representative algorithms in both multi-view depth fusion and 6D object pose estimation have ample room for improvement. This dataset will provide researchers with the necessary benchmark data to address the severe and frequent challenges that arise in manufacturing and assembly tasks when operating with highly reflective objects.

\addtolength{\textheight}{-12cm}   % This command serves to balance the column lengths

%%%%%%%%%%%%%%%%%%%%%%%%%%%%%%%%%%%%%%%%%%%%%%%%%%%%%%%%%%%%%%%%%%%%%%%%%%%%%%%%

%%%%%%%%%%%%%%%%%%%%%%%%%%%%%%%%%%%%%%%%%%%%%%%%%%%%%%%%%%%%%%%%%%%%%%%%%%%%%%%%

%%%%%%%%%%%%%%%%%%%%%%%%%%%%%%%%%%%%%%%%%%%%%%%%%%%%%%%%%%%%%%%%%%%%%%%%%%%%%%%%

%%%%%%%%%%%%%%%%%%%%%%%%%%%%%%%%%%%%%%%%%%%%%%%%%%%%%%%%%%%%%%%%%%%%%%%%%%%%%%%%

\bibliographystyle{ieeetr}
\bibliography{root.bbl}

\end{document}